\documentclass{article}

\usepackage{microtype}
\usepackage{graphicx}
\usepackage{subcaption}
\usepackage{booktabs}
\usepackage{multirow}
\usepackage{array}
\usepackage{xcolor}
\usepackage{pifont}
\usepackage{enumitem}
\newcommand{\cmark}{\ding{51}}%
\newcommand{\xmark}{\ding{55}}%

\usepackage{hyperref}

\usepackage{iftex}

\usepackage[accepted]{icml2026}

\usepackage{amsmath}
\usepackage{amssymb}
\usepackage{mathtools}
\usepackage{amsthm}

\usepackage[capitalize,noabbrev]{cleveref}

\theoremstyle{plain}

\theoremstyle{definition}

\theoremstyle{remark}

\icmltitlerunning{The Verbalization Bottleneck of CoT in Patent Claim Generation}

\begin{document}

\twocolumn[
    \icmltitle{When Reasoning Hurts Legal Drafting:\texorpdfstring{\\}{ }
    The Verbalization Bottleneck in Patent Claim Generation}

  \icmlsetsymbol{equal}{*}

  \begin{icmlauthorlist}
    \icmlauthor{Lekang Jiang}{aut1}
    \icmlauthor{Wenjun Sun}{aut1,aut2}
    \icmlauthor{Stephan Goetz}{aut1}
  \end{icmlauthorlist}

  \icmlaffiliation{aut1}{University of Cambridge}
   \icmlaffiliation{aut2}{National Science Library, Chinese Academy of Sciences; Department of Information Resources Management, School of Economics and Management, University of Chinese Academy of Sciences}

  \icmlcorrespondingauthor{Lekang Jiang}{lj408@cam.ac.uk}

  \icmlkeywords{AI for Law, Legal Reasoning, Patent Claim Generation, Chain-of-Thought, Large Language Models, Legal NLP}

  \vskip 0.3in
]

\printAffiliationsAndNotice{}

\begin{abstract}
Patent claim drafting is a challenging legal drafting task that requires technical expertise, precise linguistic control, strict adherence to formal conventions, and the preservation of complex logical relationships among claim elements. While Chain-of-Thought (CoT) prompting has been widely used to improve the reasoning capabilities of large language models (LLMs), recent evidence suggests that its benefits may be limited, or even negative, in highly structured or pattern-sensitive tasks. Therefore, this paper investigates whether CoT prompting benefits patent claim generation. We propose a task-specific CoT method for patent claim generation and evaluate its effectiveness through both automatic metrics and human expert assessment. Our results show that reasoning-enhanced prompting can improve claim quality. Moreover, we demonstrate a counter-intuitive but important empirical finding: implicit CoT, where reasoning is kept internal rather than explicitly verbalized, consistently outperforms explicit CoT. Through systematic analysis, we show that explicit CoT can introduce an unnecessary information bottleneck for claim generation. Verbalized reasoning may compromise the quality of final outputs through three specific mechanisms: abstraction of critical details, disruption of internalized generation patterns, and cascading error propagation. Our findings provide new insights into legal tasks and CoT applications.
\end{abstract}

\section{Introduction}

The automated generation of long-form, structured text, such as legal contracts, technical manuals, and patent documents, represents a significant frontier in natural language processing (NLP) \citep{zhao2023survey, jiang2025natural}. Such tasks demand not only exceptional linguistic proficiency but also stringent requirements on logical reasoning, strict adherence to structural conventions, and the integration of domain-specific knowledge. Patent claim drafting stands as a prime example of this challenge \citep{jiang2025can, wang2024autopatent}. 

Patent documents are important for protecting intellectual property (IP) and disclosing technical innovations. A central component of these documents is the claim set, which defines the legal scope of protection granted to the patent holder. Thus, writing high-quality claims is a critical task in patent drafting, which requires precise language to capture the essence of the invention while remaining legally enforceable. A patent claim must articulate the core novelty of an invention using hierarchical, logically rigorous, and legally precise language \citep{epo2020}. The complexity of this process is further compounded by its long contexts with over 1,000 tokens on average \citep{suzgun2024harvard}. The challenges make the manual drafting time-consuming and intellectually intensive even for human experts. Thus, the automation of the patent drafting process, especially for claims, is important. 

Despite the advancements of large language models (LLMs) in general-purpose text generation, the application to highly structured and complex tasks such as patent drafting often shows suboptimal results, characterized by logical imprecision, structural inconsistencies, or deviation from core requirements \citep{jiang2025natural}. To unlock and guide the deeper reasoning capabilities of LLMs, the Chain-of-Thought (CoT) prompting technique has emerged as a powerful method \citep{wei2022chain, kojima2022large, wangself, zhangautomatic}. CoT encourages models to break down complex problems into a series of intermediate reasoning steps before arriving at a final answer. In consequence, it could significantly enhance performance on arithmetic, commonsense, and symbolic reasoning tasks. 

However, recent studies have reported that the effectiveness of CoT in some tasks, such as pattern-based and patent-related tasks, is limited or even counterproductive \citep{zheng2025curse, wang2025ipbench}. These findings raise critical questions about whether CoT can benefit patent claim drafting. Overall, we make the following contributions.

(1) We propose a task-specific zero-shot CoT prompting method for patent claim generation and demonstrate its effectiveness through both automated and human expert evaluations.

(2) We demonstrate a counter-intuitive but important empirical finding: implicit CoT (LLMs are guided to think internally and only output the final claims) consistently outperforms explicit CoT (intermediate reasoning steps are explicitly written out prior to claim generation).

(3) Through diagnostic experiments and case studies, we trace the decrease to an information bottleneck created by reasoning externalization, which manifests through three mechanisms: (1) verbalized reasoning abstracts away legally critical details; (2) the explicit reasoning phase disrupts the model's internalized generation patterns; and (3) errors cascade irrecoverably across reasoning steps.

\section{Related Work}
\label{sec:related}

\subsection{CoT Prompting}
\citet{wei2022chain} introduced CoT prompting, which provides LLMs with few-shot exemplars that include intermediate reasoning steps to improve the models' performance on complex arithmetic, commonsense, and symbolic reasoning tasks. This finding suggested that eliciting a step-by-step reasoning process is key to unlocking the latent capabilities of LLMs. Subsequent work by \citet{kojima2022large} revealed that this reasoning ability could be unlocked even without few-shot exemplars. Their zero-shot CoT approach merely appends a simple, generic instruction, such as ``Let's think step by step'', to the prompt. This discovery was pivotal, as it showed that the reasoning mechanism is an emergent capability of sufficiently scaled models, which can be triggered by a direct instruction rather than imitation. More complex reasoning structures have also been explored, such as the Tree of Thoughts (ToT) \citep{yao2023tree} and the Graph of Thoughts (GoT) \citep{besta2024graph}. However, recent studies have reported that the effectiveness of CoT in some tasks, including patent-related tasks, is limited or even counterproductive \citep{zheng2025curse, wang2025ipbench}.

\subsection{Patent Claim Generation}
Drafting high-quality patent claims is a time-consuming, costly, and skill-intensive process \citep{epo2020}. Thus, recent research has explored LLMs to assist inventors and patent professionals in automatically generating well-structured claims \citep{jiang2025natural}. The research focus has gradually shifted from earlier approaches that train neural models on patent abstracts \citep{lee2020patentgenerate, lee2020controlling} towards using LLMs to generate claims from detailed patent descriptions \citep{jiang2025can, wang2024autopatent} or revise draft claims \citep{jiang-etal-2025-patent}. As descriptions provide richer technical context, these methods can significantly improve the completeness and precision of the generated claims. More recent works start to investigate optimization methods, such as using agentic frameworks and multimodal inputs \citep{shea2025autospec, yang2026patentvision,yu2026toc}. This work explores whether CoT can benefit patent claim drafting.

\section{Task and Method}
\label{sec:method}

\subsection{Task Formulation}

The task aims to automatically generate a set of high-quality patent claims $C$ based on a given technical description $D$ of an invention: $C = G(I, D)$, where $G$ represents LLMs, and $I$ is the instruction prompt. 
Unlike conventional text generation tasks such as summarization or translation, generating patent claims presents a significant challenge that demands deep semantic understanding and complex logical reasoning. The model must distill the core inventive concepts from a lengthy technical document and synthesize these concepts into a hierarchically structured set of claims. Each dependent claim must logically narrow the scope of its parent claim and adhere to strict legal formalism. Furthermore, every element within the claims must be fully supported by and grounded in the provided description. Therefore, an intermediate reasoning process is recommended before writing final claims to identify essential invention features, structure features logically, and articulate claims formally. 

\subsection{Zero-Shot CoT Prompting}

\begin{figure}[!t]
    \centering
    \includegraphics[width=\columnwidth]{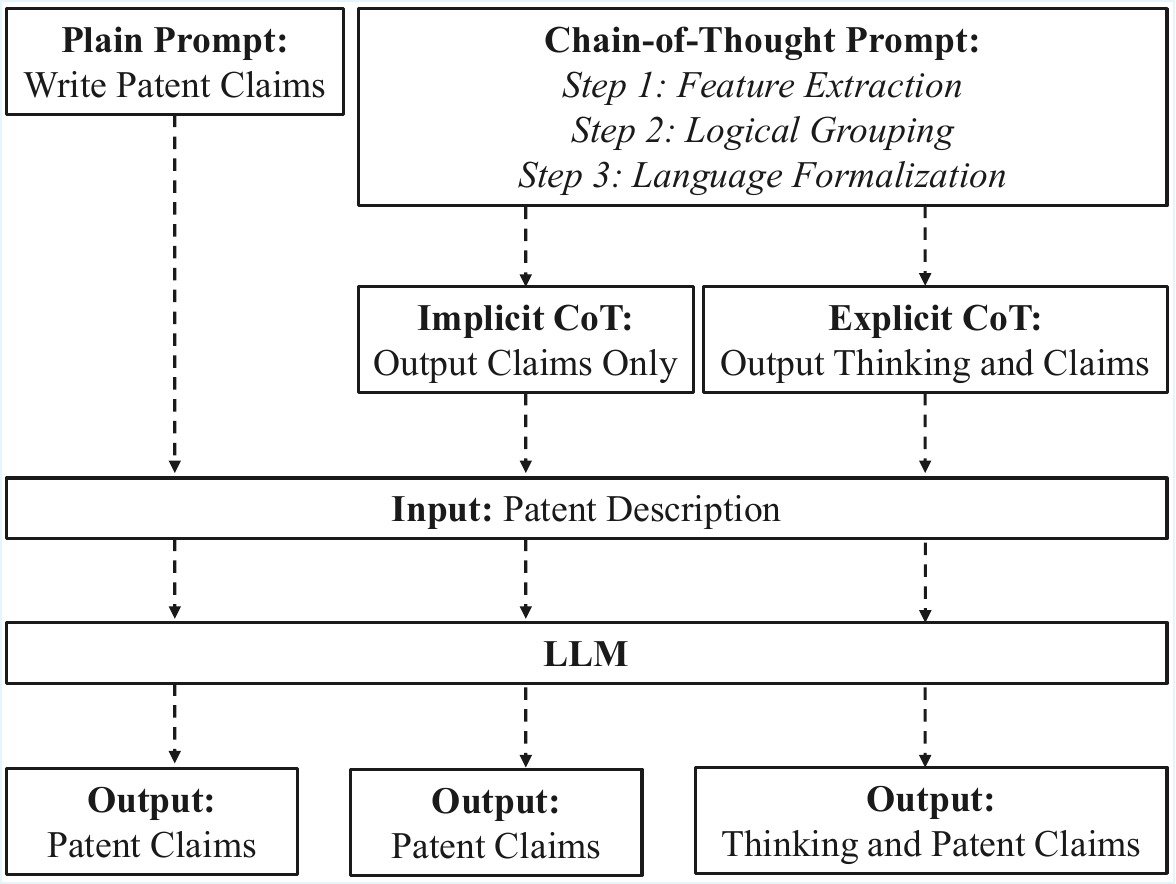}
    \caption{Overview of the CoT prompting methods. Implicit CoT and Explicit CoT share the same three-step process (Feature Extraction $\to$ Logical Grouping $\to$ Language Formalization); they differ in whether intermediate reasoning is externalized as text.}
    \label{fig:overview}
\end{figure}

Although CoT has significantly advanced the general reasoning capabilities of LLMs, it often relies on either manually crafted exemplars \citep{wei2022chain} or generic instructions \citep{kojima2022large} that may not be optimal for highly specialized, domain-specific tasks that demand a precise logical flow. Our work builds upon the principles of zero-shot CoT \citep{kojima2022large} and addresses the domain gap. In contrast to the use of a generic trigger phrase, we introduce a structured zero-shot CoT approach: $C = G(I_{CoT}, D)$ as shown in Figure \ref{fig:overview}. Our approach decomposes the generation process into three structured steps to simulate the professional drafting process. The three-stage decomposition is motivated by the functional requirements of patent claim drafting. In particular, claim drafting requires the identification of invention-defining features, the organization of those features into legally and technically coherent dependencies, and their expression using conventional claim language. These requirements motivate the stages of feature extraction, logical grouping, and language formalization, respectively. Completed CoT prompts are illustrated in Appendix Table \ref{tab:prompt}. 

\textbf{Step 1: Feature Extraction. }
The model first identifies the key technical features in patent descriptions and their importance ranking. This step encourages the model to recognize essential inventive aspects to improve the content completeness in the final claims. 

\textbf{Step 2: Logical Grouping. }
The extracted features are organized logically based on their functional relationships and dependencies. This step supports the structured composition of features to enhance the logical structure of the final claims.

\textbf{Step 3: Language Formalization. }
Organized features are written into textual claims using formal patent language to ensure clarity, consistency, and technical accuracy.

By providing the model with a task-specific framework via structured instructions, we guide its reasoning process to align with the domain-specific requirements of patent claim drafting. This method maintains the simplicity of the zero-shot approach but imposes a necessary structure to enhance the completeness, logical coherence, and language formality of the generated output.

\subsection{CoT Variants}
We experiment with the plain prompt and different CoT prompting variants (Figure \ref{fig:overview}):

\noindent \textbf{Plain Prompt}: the model outputs only the final claims without three-step CoT prompting: $C = G(I, D)$.

\noindent \textbf{Implicit CoT}: the model is guided to think step-by-step internally and only outputs the final claims: $C = G(I_{CoT\_Implicit}, D)$.

\noindent \textbf{Explicit CoT}: the intermediate reasoning steps are explicitly written out prior to claim generation: $R, C = G(I_{CoT\_Explicit}, D)$, where $R$ is the reasoning process.

\section{Experimental Setup}
\label{sec:setup}

\subsection{Datasets}

We conduct main experiments on the EPD dataset \citep{jiang2025enriching}, which contains 693 English-language patents from the European Patent Office (EPO) for claim generation. We choose this dataset for two main reasons: (1) EPD features high-quality granted patents, in contrast to the application versions used in prior work \citep{jiang2025can}; and (2) EPD better reflects real-world scenarios, which enables a more rigorous evaluation of LLM performance. For generalization exploration, we also test the method on the HUPD-DCG dataset \citep{jiang2025can}, with 1,311 patents from the United States Patent and Trademark Office (USPTO), and find similar results. Results are consistent with those on EPD and are presented in Appendix~\ref{app:moreresults}.

\subsection{Models}

We mainly focus on Llama-3.1-8B \citep{dubey2024llama}, Mixtral-8$\times$7B \citep{jiang2024mixtral}, and GPT-4o \citep{openai2024gpt4o} in the paper to explore the effect of model sizes and architectures. We further report the results of Mistral-7B \citep{jiang2023mistral} and Llama-3.1-70B \citep{dubey2024llama} in Appendix \ref{app:moreresults}.

\begin{table*}[ht!]
\footnotesize
\centering
\caption{Automated and human expert evaluation results for different prompting methods. Human-evaluation entries are reported as mean ± standard deviation across the evaluated samples. }
\resizebox{\linewidth}{!}{
\begin{tabular}{l|c|cccc|cccc}
\toprule
& \multicolumn{1}{c|}{\textbf{PatClaimEval}}  
& \multicolumn{4}{c|}{\textbf{LLM-as-a-Judge}} 
& \multicolumn{4}{c}{\textbf{Human Evaluation}} \\
\cmidrule(lr){2-2} \cmidrule(lr){3-6} \cmidrule(lr){7-10}
\textbf{Model / Setting} 
& \textbf{Quality} 
& \textbf{Content} & \textbf{Logic} & \textbf{Language} & \textbf{Average} 
& \textbf{Content} & \textbf{Logic} & \textbf{Language} & \textbf{Average}  \\
\midrule

Llama-3.1-8B & 21.8 & 73.1 & 68.7 & 69.4 & 70.2 & 82.3$\pm$2.4 & 82.5$\pm$2.3 &	81.2$\pm$2.4 & 82.0 \\
\quad \textit{w/ Implicit CoT} & 21.9 & 74.1 & 69.5 & 69.4 & 70.7 & 82.7$\pm$2.2 &	82.5$\pm$2.1 & 81.8$\pm$2.1 & 82.3 \\
\quad \textit{w/ Explicit CoT} & 21.7 & 59.5 & 57.9 & 60.9 & 59.1 & 82.4$\pm$2.2 & 82.5$\pm$2.3 & 81.4$\pm$2.0 & 82.1 \\
\midrule

Mixtral-8$\times$7B & 20.6 & 75.6 & 73.0 & 71.8  & 73.2 & 84.6$\pm$2.3 & 85.0$\pm$2.4 & 84.6$\pm$2.1 & 84.7  \\
\quad \textit{w/ Implicit CoT} & 20.3 & 74.9 & 72.9 & 73.1 & 73.3 & 85.4$\pm$2.0 & 85.4$\pm$2.1 & 85.8$\pm$2.0 & 85.5 \\
\quad \textit{w/ Explicit CoT} & 20.9 & 68.0 & 67.3 & 68.4 & 67.5 & 84.6$\pm$2.2 & 85.2$\pm$2.2 & 84.7$\pm$2.1 & 84.8 \\

\midrule

GPT-4o & 23.5 & 77.3 & 76.4 & 74.8 & 75.9 & 88.5$\pm$2.4 & 86.3$\pm$2.5 & 87.7$\pm$1.8 & 87.5  \\
\quad \textit{w/ Implicit CoT} & 23.7 & 78.0 & 76.8 & 75.4 & 76.5 & 93.0$\pm$1.5 & 92.2$\pm$2.1 & 92.8$\pm$1.6 & 92.7 \\
\quad \textit{w/ Explicit CoT} & 22.5 & 70.3 & 70.6 & 71.3 & 70.4 & 90.0$\pm$1.8 & 89.8$\pm$2.3 & 91.7$\pm$1.6 & 90.5 \\
\bottomrule
\end{tabular}

}

\label{table:main}
\end{table*}

\subsection{Evaluation Protocol}

\paragraph{Human Expert Evaluation.}
We invite licensed patent attorneys to evaluate generated claims compared to the gold claims in three aspects (scores 0--100): (1) Content Coverage: The extent to which the draft claims encapsulate all critical technical aspects of the invention. (2) Logical Structure: The degree to which the claim set exhibits a coherent, well-organized, and legally sound logical structure, including appropriate dependencies and hierarchical relationships. (3) Language Quality: The clarity, precision, and formal appropriateness of the language used in the claim set, including terminology, syntax, and adherence to patent drafting conventions. Patent attorneys are provided with reference claims (granted versions) and candidate claims generated by different models. The instructions are: \textit{You are given a list of patent claims generated by different models. Compare them with the gold claim (granted version) and assess their quality.} Since the evaluation process is time-intensive and costly, we randomly select 120 samples for human evaluations, and each example is rated by one patent expert.

\paragraph{Automated Evaluation.}
We use PatClaimEval \citep{jiang2025towards}, which is specifically designed for patent claim evaluation. Moreover, studies have shown that LLM-as-a-judge evaluators can achieve better human alignment, including patent claim evaluation \citep{liu2023g, jiang2025towards}. To alleviate potential bias, we use the Claude-Opus-4.6 to evaluate generated patent claims based on the above three human evaluation dimensions. Detailed settings and prompts are reported in Appendix~\ref{llmasjudge}. 

\paragraph{Qualitative Analysis.} 
We provide concrete model outputs and analysis to intuitively demonstrate the performance differences among various prompting strategies. Our analysis focuses on a side-by-side comparison of claims generated from the same patent description to highlight the typical failure modes of baselines and showcase the advantages of CoT methods. 

\paragraph{Stepwise Diagnostic.}
For \emph{Explicit CoT}, we additionally annotate each reasoning stage to locate where degradation originates. We randomly select 80 examples for human evaluation. For feature extraction, we compare the list of features the model identifies in its thinking block against the list distilled from the gold claims. Then, we calculate the precision, recall, and F1 scores. For logical grouping, feature usage rate is measured by the number of features used in the grouping step divided by the total number of features extracted. We also rate the logical correctness in the second step, ranging from 0 to 100. For step three, the faithfulness score is assessed on a 100-point scale, which measures whether the final claim is a direct and logical product of the preceding reasoning. A high score indicates the final claim is a perfect synthesis of the reasoning plan, where all key elements are traceable.

\begin{table*}[t]
\footnotesize
\centering
\caption{Intermediate-step human analysis for \emph{Explicit CoT} prompting. }
\label{table:steps}
\begin{tabular}{lcccccc}
\toprule
& \multicolumn{3}{c}{\textbf{Step 1: Extraction}} & \multicolumn{2}{c}{\textbf{Step 2: Grouping}} & \multicolumn{1}{c}{\textbf{Step 3: Formalization}}\\
\cmidrule(lr){2-4} \cmidrule(lr){5-6} \cmidrule(lr){7-7}
\textbf{Model} & \textbf{Precision} & \textbf{Recall} & \textbf{F1} & \textbf{Usage} & \textbf{Correctness} & \textbf{Faithfulness} \\
\midrule
Llama-3.1-8B & 62 & 56 & 58 & 98 & 65 & 70  \\
Mixtral-8$\times$7B & 70 & 53 & 60 & 99 & 69 & 73  \\
GPT-4o & 72 & 58 & 64 & 99 & 72 & 77 \\
\bottomrule
\end{tabular}
\end{table*}

\section{Main Results}
\label{sec:results}

Table~\ref{table:main} presents both automated and human expert evaluation results across different models. For qualitative analysis, we provide a representative case study in Table~\ref{tab:feature_comparison} and include full model outputs in Table~\ref{table:example_ouput}. We mainly rely on human expert evaluation results and use automated evaluation as supplementary evidence. 

\subsection{CoT Empowers Patent Claim Generation}

Our results show that implicit CoT prompting significantly improves claim generation quality. While we agree that the gains for smaller models are modest, the improvement for larger models such as GPT-4o is both substantial and stable, with higher evaluation scores and smaller standard deviations. As shown in Table \ref{table:main}, GPT-4o with implicit CoT achieves an average increase of 5.2 points in human evaluations, with the most notable gain observed in logical structure (5.9 points). Similarly, automated evaluations reveal consistent improvements of GPT-4o, including a 0.2-point increase on PatClaimEval (23.5 to 23.7) and an average 0.6-point increase using LLM-as-a-judge (75.9 to 76.5).

An examination of the example outputs in Table~\ref{tab:feature_comparison} and \ref{table:example_ouput} reveals that GPT-4o with implicit CoT demonstrates higher content and structure similarity to the gold claim compared to the standard prompt. The core inventive concept of the gold claim is the water reservoir's construction, specifically the use of ``blind holes'' or ``non-smooth textures'' on the interior surfaces of the tray. Although the standard prompt correctly identifies the invention's general purpose as a humidified container for an intraocular lens, it fails to capture this critical structural detail. In contrast, the implicit CoT approach accurately and explicitly reproduces this key feature logically.

These improvements can be attributed to two key factors. First, CoT decomposes claim generation into three sub-tasks that clarify task objectives at each stage, reduce cognitive load, and guide models to generate better results. Second, this decomposition activates the model's latent reasoning, planning, and domain knowledge capabilities, which may remain underutilized in direct generation. 

\subsection{Implicit CoT outperforms Explicit CoT}

We observe that explicit CoT substantially degrades performance compared to implicit CoT across both human and automated evaluations. As shown in Table \ref{table:main}, explicit CoT for GPT-4o leads to consistent drops across all human evaluation dimensions, with an average decrease of 2.2 points. Similarly, Table~\ref{table:main} reports notable declines in automated metrics for GPT-4o, including a 1.2-point drop on PatClaimEval (23.7 to 22.5) and an average 6.1-point reduction when using LLM-as-judge (76.5 to 70.4). The decreases in other models are even worse. For example, the average LLM-as-a-judge performances of explicit CoT are more than 10 points lower than implicit CoT for Llama-3.1-8B. This result contrasts with prior findings in arithmetic or logic reasoning tasks, where explicit CoT typically improves model accuracy \citep{wei2022chain,kojima2022large, yu2024llms}.

\begin{table*}[t]
\centering
\footnotesize
\caption{Comparison of technical features of GPT-4o's outputs on patent EP3060168B1. \cmark~means correctly stated, and \xmark~means not mentioned or inconsistent with the gold claim. Detailed model outputs are in Appendix Table~\ref{table:example_ouput}.}
\label{tab:feature_comparison}
\resizebox{\textwidth}{!}{
\begin{tabular}{l|c|c|c|cc}
\toprule
 & \textbf{Gold} & \textbf{Plain Prompt} & \textbf{Implicit CoT} & \multicolumn{2}{c}{\textbf{Explicit CoT}} \\
\cmidrule(lr){2-6}
\textbf{Features} & \textbf{Claims} & \textbf{Claims} &  \textbf{Claims} & \textbf{Thinking} & \textbf{Claims} \\
\midrule
\multicolumn{6}{l}{\textit{\textbf{Core Components}}} \\
\quad Air-tight container, Injector, Intraocular Lens (IOL), Water reservoir  & \cmark & \cmark & \cmark & \cmark & \cmark \\
\midrule
\multicolumn{6}{l}{\textit{\textbf{Container Details}}} \\
\quad Comprises a tray, Side walls, Bottom walls          & \cmark & \xmark & \cmark & \xmark & \cmark \\
\midrule
\multicolumn{6}{l}{\textit{\textbf{IOL Details}}} \\
\quad Foldable   & \cmark & \cmark & \cmark & \cmark & \cmark \\
\quad Hydrophilic               & \cmark & \xmark & \xmark & \xmark & \xmark \\
\quad Unfolded in injector      & \cmark & \xmark & \cmark & \xmark & \cmark  \\
\midrule
\multicolumn{6}{l}{\textit{\textbf{Water Reservoir Details}}} \\
\quad Provided by blind holes    & \cmark & \cmark & \cmark & \xmark & \cmark  \\
\quad Provided by non-smooth textures & \cmark & \xmark & \cmark & \xmark & \xmark \\
\quad Specific water volume (0.5--3 ml) & \cmark & \xmark & \xmark & \xmark & \xmark \\
\bottomrule
\end{tabular}
}
\end{table*}

\section{The Verbalization Bottleneck}
\label{sec:bottleneck}

The consistent underperformance of explicit CoT demands explanation. We argue that the root cause lies in a fundamental mismatch between what explicit CoT does and what patent claim generation requires. In reasoning tasks such as arithmetic, intermediate steps can be expressed in natural language with minimal information loss---each step is a self-contained, verifiable proposition. Patent claim generation is qualitatively different: the ``reasoning'' involved---selecting which technical details are legally critical, weighing alternative phrasings, planning hierarchical claim structure---is deeply entangled and hard to be faithfully linearized into text. When the model is forced to externalize this process, the resulting reasoning text acts as an \emph{information bottleneck}: the model's rich internal representations are compressed into a sequential, lossy verbal summary, and all subsequent generations are conditioned on this degraded intermediate rather than the original patent description. 

Through detailed examination of model outputs, we conclude that the information bottleneck manifests through three concrete mechanisms.

\subsection{Mechanism 1: Abstraction of Critical Detail}
\label{sec:mech1}

Patent claims demand specificity: legally enforceable protection depends on concrete technical features, exact structural configurations, and precise numerical parameters---not high-level summaries. However, explicit CoT's reasoning stage systematically abstracts away the detail, which replaces specific component names and structural relationships with vague, higher-level notions. This abstraction tendency is well-suited for reasoning tasks, where generalizing from specifics is precisely what enables multi-step deduction~\citep{wei2022chain}. For patent claims, however, abstraction is destructive: the task requires preserving detail through generation, not reducing it through reasoning. The information bottleneck thus imposes a systematic bias toward the wrong kind of transformation, which breaks the specificity that patent claims depends on.

\subsection{Mechanism 2: Disruption of Internalized Generation Patterns}
\label{sec:mech2}

Through pretraining on large corpora of legal and technical documents, LLMs acquire robust internalized patterns for generating structured text---including appropriate terminology, hierarchical claim organization, and consistent cross-referencing between claim elements. Inserting an explicit reasoning phase before generation appears to disrupt these patterns. The verbalized reasoning, rather than enriching the generation process, interposes a fragmentary and incomplete conceptual plan that overrides the model's capacity to produce structurally coherent output directly from the source description.
 
Our findings are consistent with recent work showing that explicit CoT can be detrimental in certain tasks~\citep{ma2025reasoning, wang2025ipbench}. For example, explicit CoT has been shown to introduce noisy and redundant reasoning traces \citep{zheng2025curse}, reduce instruction-following accuracy \citep{li2025thinking}, and produce explanations that are misaligned with the model’s internal decision process \citep{turpin2023language}. Collectively, these observations provide strong empirical support for our findings.

\subsection{Mechanism 3: Error Cascade Across Reasoning Steps}
\label{sec:mech3}

To directly assess the quality of intermediate reasoning, we conduct a diagnostic human evaluation of each explicit CoT stage (Table~\ref{table:steps}). The results reveal a clear error cascade: Step~1 (feature extraction) achieves only 62--72\% precision across models, which means essential technical features are missed. Step~2 (logical grouping) propagates and amplifies these omissions, which scores 65--72\% in logical correctness. Step~3 (language formalization) generates claims conditioned on an inaccurate reasoning trajectory and imperfect adherence to the intermediate plan, which leads to compounded degradation. These errors are irrecoverable by design: once the verbalized reasoning omits a critical feature, no subsequent step can restore it, because the model generates conditioned on the intermediate text rather than the full patent description. Implicit CoT avoids this cascade entirely by keeping intermediate representations internal, where they remain rich and uncompressed.

\subsection{Case Study}
\label{sec:case}

Table~\ref{tab:feature_comparison} presents a detailed comparison of model outputs to support the above analysis. Explicit CoT's thinking stage abstracts ``blind holes'' into generic notions like ``water reservoir systems'' and ``distribution enhancers,'' which loses the structural specificity essential for legal protection (Mechanism~1). It also omits structural elements such as ``tray'' and ``side walls'' from the reasoning plan, and these omissions propagate directly into the final claims (Mechanism~3). Notably, implicit CoT correctly captures both ``blind holes'' and the tray structure---features that the model clearly has the capacity to identify but fails to preserve when forced through the verbalization bottleneck (Mechanism~2). The full model outputs in Appendix Table~\ref{table:example_ouput} confirm that the explicit CoT claims follow a flawed blueprint established during reasoning, while implicit CoT produces claims with substantially higher fidelity to the gold claim.

\section{Conclusion and Discussion}
\label{sec:discussion}

This paper investigates when reasoning helps or hurts legal drafting. In the context of patent claim generation, we found that CoT prompting is beneficial when used as silent planning, but becomes less reliable when the same reasoning is explicitly verbalized before drafting. This pattern holds across different patent datasets, model families, and evaluation metrics. 
We interpret this result as a \emph{verbalization bottleneck}. Patent claims encode legal effect through precise wording, element-level dependencies, and drafting conventions. When intermediate reasoning is forced into visible natural language, it can compress legally details into an abstract plan, disturb the model's internalized drafting routines, and propagate early reasoning errors into the final claim.

\paragraph{Beyond patents.}
Although our experiments focus on patent claims, the verbalization-bottleneck hypothesis may extend to other detail-preserving legal drafting tasks, such as contract clause generation, regulatory filing preparation, compliance documentation, and statutory text generation. These settings similarly assign legal significance to concrete wording, dependencies, exceptions, and formal structure. The present study therefore offers both a methodological template and a set of diagnostic mechanisms for future AI-for-law evaluation: vary the visibility of reasoning while holding the underlying scaffold fixed, and examine whether verbalization changes the quality of the resulting legal artifact.

\paragraph{Limitations.}
This study has several limitations. First, we focus on zero-shot prompting; few-shot prompting, supervised fine-tuning, retrieval-augmented generation, or agentic drafting workflows may interact differently with the verbalization bottleneck. Second, we do not perform hyperparameter tuning during inference; instead, we keep the hyperparameters fixed across all experiments to ensure fair comparisons. Third, our bottleneck account is explanatory rather than a formal information-theoretic proof. Fourth, our mechanism analysis relies on expert annotation, stepwise diagnostics, and case studies rather than representation-level causal intervention. Overall, these limitations suggest that the verbalization bottleneck should be treated not as a settled property of all legal reasoning tasks, but as a testable hypothesis for structured legal generation: in some legal settings, making reasoning visible may improve transparency while reducing drafting fidelity. In addition, reinforcement learning may be useful for patent reasoning tasks \citep{jiang-etal-2026-reasoning}.

\bibliographystyle{icml2026}
\bibliography{main}


\appendix

\begin{table*}[ht!]
\footnotesize
\centering
\caption{Automated evaluation results of different models on EPD dataset.  }
\label{table:aux}

\begin{tabular}{lcccccccc}
\toprule
& \multicolumn{3}{c}{\textbf{Traditional Metrics}} & \multicolumn{1}{c}{\textbf{PatClaimEval}}  & \multicolumn{4}{c}{\textbf{LLM-as-a-Judge}} \\
\cmidrule(lr){2-4} \cmidrule(lr){5-5} \cmidrule(lr){6-9}
\textbf{Model} & \textbf{BLEU} & \textbf{R-L} & \textbf{BERTScore} & \textbf{Quality} & \textbf{Content} & \textbf{Logic} & \textbf{Language} & \textbf{Average}  \\
\midrule

Mistral-7B & 31.4 & 41.3 & 85.6 & 17.6 & 72.8 & 68.9 & 68.9 &  69.9 \\
\quad \textit{w/ Implicit CoT} & 31.0 & 40.1 & 85.5 & 18.1 & 73.4 & 69.5 & 68.3 & 70.1 \\
\quad \textit{w/ Explicit CoT} & 15.5 & 33.0 & 83.9 & 16.3 & 63.2 & 60.3 & 56.8 & 59.9 \\
\midrule

Llama-3.1-70B & 27.2 & 38.9 & 85.6 & 20.9 & 75.9 & 74.2 & 74.3 & 74.5 \\
\quad \textit{w/ Implicit CoT} & 26.3 & 38.2 & 85.6 & 21.1 & 75.5 & 75.0 & 74.3 & 74.5 \\
\quad \textit{w/ Explicit CoT} & 11.7 & 35.9 & 84.7 & 22.0 & 61.0 & 64.0 & 65.0 & 63.1 \\

\bottomrule
\end{tabular}

\end{table*}

\begin{table*}[ht!]
\footnotesize
\centering
\caption{Automated evaluation results of different models on the HPUD-DCG dataset.}
\label{table:main_us}

\begin{tabular}{lcccccccc}
\toprule
& \multicolumn{3}{c}{\textbf{Traditional Metrics}} & \multicolumn{1}{c}{\textbf{PatClaimEval}}  & \multicolumn{4}{c}{\textbf{LLM-as-a-Judge}} \\
\cmidrule(lr){2-4} \cmidrule(lr){5-5} \cmidrule(lr){6-9}
\textbf{Model} & \textbf{BLEU} & \textbf{R-L} & \textbf{BERTScore} & \textbf{Quality} & \textbf{Content} & \textbf{Logic} & \textbf{Language} & \textbf{Average}  \\
\midrule

Mistral-7B  & 36.0 & 40.6 & 87.0 & 16.5 & 71.4 & 71.5 & 69.4 & 70.5 \\
\quad \textit{w/ Implicit CoT} & 35.6 & 40.9 & 86.9 & 18.4 & 72.3 & 72.9 & 70.9 & 71.7 \\
\quad \textit{w/ Explicit CoT} & 19.6 & 33.0 & 85.1 & 18.3 & 62.7 & 62.2  & 60.3 & 61.5 \\
\midrule

GPT-4o & 22.3 & 39.0 & 86.5 & 22.0 & 73.2 & 73.3 & 72.2 & 72.6 \\
\quad \textit{w/ Implicit CoT} & 20.4 & 37.6 & 86.4 & 23.0 & 73.4 & 74.1 & 73.3 & 73.4 \\
\quad \textit{w/ Explicit CoT}  & 13.5 & 33.0 & 86.0 & 22.4 & 65.8 & 71.0 & 71.3 & 69.0 \\

\bottomrule
\end{tabular}

\end{table*}
\begin{table*}[ht!]
\footnotesize
\centering
\caption{Automated and human evaluation results of ablation study on GPT-4o. }
\label{table:abl}
\resizebox{0.99\linewidth}{!}{
\begin{tabular}{l|c|cccc|cccc}
\toprule
& \multicolumn{1}{c|}{\textbf{PatClaimEval}}  
& \multicolumn{4}{c|}{\textbf{LLM-as-a-Judge}} 
& \multicolumn{4}{c}{\textbf{Human Evaluation}} \\
\cmidrule(lr){2-2} \cmidrule(lr){3-6} \cmidrule(lr){7-10}
\textbf{Model / Setting} 
& \textbf{Quality} 
& \textbf{Content} & \textbf{Logic} & \textbf{Language} & \textbf{Average} 
& \textbf{Content} & \textbf{Logic} & \textbf{Language} & \textbf{Average}  \\
\midrule

GPT-4o & 23.5 & 77.3 & 76.4 & 74.8 & 75.9 & 88.5 & 86.3 & 87.7 & 87.5  \\
\quad \textit{w/ Implicit CoT} & 23.7 & 78.0 & 76.8 & 75.4 & 76.5 & 93.0 & 92.2 & 92.8 & 92.7 \\
\quad \quad \textit{w/o Extraction}& 23.5 & 78.1 & 76.4 & 75.0 & 76.2 & 93.3 & 92.7 & 93.2 & 93.1 \\
\quad \quad \textit{w/o Grouping} & 23.4 & 77.0 & 75.5 & 74.4 & 75.4 & 91.7 & 90.7 & 92.2 & 91.5  \\
\quad \quad \textit{w/o Formalization} & 24.3 & 77.2 & 75.0 & 74.1 & 75.2 & 91.3 &	91.8 &	91.8 &	91.7 
\\

\bottomrule
\end{tabular}
}

\end{table*}

\section{Experimental Details}
\label{app:experimentdetails}

Detailed prompts for claim generation are reported in Table \ref{tab:prompt}. The inference processes of Llama\footnote{\url{https://huggingface.co/meta-llama/Llama-3.1-8B-Instruct}, \url{https://huggingface.co/meta-llama/Llama-3.1-70B-Instruct}}, Mistral\footnote{\url{https://huggingface.co/mistralai/Mistral-7B-Instruct-v0.1}}, and Mixtral\footnote{\url{https://huggingface.co/mistralai/Mixtral-8x7B-Instruct-v0.1}}are conducted on NVIDIA A100 GPUs. The total running time is about 20 hours. Inference of GPT-4o\footnote{GPT-4o: \url{https://platform.openai.com/docs/models/gpt-4o}} is through its API. The hyperparameters we used are: temperature of 0.1, top\_p of 0.95, and maximum output tokens of 4096. We adopt traditional metrics for text generation, including BLEU \citep{papineni2002bleu}, ROUGE-L (R-L) \citep{lin2004rouge}, and BERTScore \citep{zhang2019bertscore}, to assess surface-level sequence overlaps and semantic similarity. We use the package from the HuggingFace \textit{evaluate} library.\footnote{\url{https://github.com/huggingface/evaluate}}

\section{LLM-as-a-Judge Prompt}
\label{llmasjudge}

We use the following prompt for LLM-as-a-judge evaluations, as shown in Table~\ref{tab:geval}. The hyperparameters we used are: temperature of 0.1, top\_p of 0.95, and maximum output tokens of 128. To alleviate potential bias, we use Claude-Opus-4.6 \footnote{\url{https://www.anthropic.com/news/claude-opus-4-6}} for evaluation.

\section{More Results}
\label{app:moreresults}

\subsection{Evaluation on More Models}

We report the automated evaluation results of Mistral-7B and Llama-3.1-70B on the EPD dataset in Table \ref{table:aux}.

\begin{table*}[!t]
\centering
\footnotesize
\caption{LLM-as-a-judge prompt used for claim evaluation. }
\label{tab:geval}

\begin{tabular}{|p{0.95\linewidth}|}
\toprule
\textbf{Instructions:} \\
You will be given the draft claims and the referenced claims of the same patent. Your task is to rate the draft claims on three metrics using the referenced claims as the gold standard. Please make sure you read and understand these instructions carefully. Please keep this document open while reviewing, and refer to it as needed. \\
\\
\textbf{Evaluation Criteria:} \\
\textbf{1. Content Coverage (0--100):}\\
 The extent to which the draft claims encapsulated all critical technical aspects of the invention. \\
\quad 0--20: Most essential features are missing or poorly described. \\
\quad 21--40: Some essential features are present but significant gaps remain.\\
\quad 41--60: Majority of essential features are covered but with minor omissions. \\
\quad 61--80: Almost all essential features are well described with very few gaps. \\
\quad 81--100: All essential features are thoroughly and comprehensively covered. \\
\\
\textbf{2. Logical Structure (0--100):} \\
The degree to which the claim set exhibits a coherent, well-organized, and legally sound logical structure, including appropriate dependencies and hierarchical relationships. \\
\quad 0--20: Claim structure is confusing or non-compliant; dependencies are incorrect or illogical. \\
\quad 21--40: Some structure is evident but many dependency relationships are unclear, misplaced, or violate drafting norms. \\
\quad 41--60: Basic structure is acceptable; most dependencies are logical though some organizational issues persist.  \\
\quad 61--80: Generally well-structured with minor issues in hierarchy or ordering; dependencies mostly valid.  \\
\quad 81--100: Excellent structural clarity with fully compliant, logical, and easily navigable claim hierarchy and dependencies.  \\
\\
\textbf{3. Language Quality (0--100):}\\
The clarity, precision, and formal appropriateness of the language used in the claim set, including terminology, syntax, and adherence to patent drafting conventions.  \\
\quad 0--20: Language is vague, ambiguous, or ungrammatical; terminology is inconsistent or incorrect. \\
\quad 21--40: Basic meaning is somewhat understandable but style is inconsistent, with notable issues in clarity or precision. \\
\quad 41--60: Mostly clear and accurate, though with occasional awkward phrasing or imprecise terminology. \\
\quad 61--80: Generally polished and professional language with only minor lapses in clarity or consistency. \\
\quad 81--100: Language is highly precise, concise, and consistent, following patent norms and clearly conveying the technical content. \\
\\

\textbf{Evaluation Steps:} \\
1. Read the referenced claims carefully and assume the referenced claims have scores of 100 in all Evaluation Criteria. \\
2. Read the draft claims and compare it to the referenced claims. \\
3. Assign a score for each metric based on the Evaluation Criteria. \\
\\
\textbf{Example:} \\
Referenced Claims: \texttt{<<Claims>>} \\
Draft Claims: \texttt{<<Claims>>} \\
Evaluation Form (scores ONLY): \\
- Content Coverage: X, \\
- Logical Structure: X, \\
- Language Quality: X. \\

\bottomrule
\end{tabular}

\end{table*}

\subsection{Results on HUPD Dataset}

We report the automated evaluation results on the HUPD-DCG dataset in Table \ref{table:main_us}.

\subsection{Ablation Study}

We conduct ablation studies to assess the effectiveness of different stages in CoT prompting. \textbf{Removing either the logical grouping or the language formalization step leads to clear performance degradation in both human and automated evaluations.} As shown in Table \ref{table:abl}, these variants result in over 1-point average declines, with the largest drop observed in the logical structure dimension (over 1.5 points) when the logical grouping step is removed. These results underscore the critical role of both steps, especially logical grouping, in generating high-quality patent claims.

\textbf{Interestingly, removing the feature extraction step slightly improves human evaluation scores.} Explicitly prompting the model to extract features may fragment its focus during the more critical logical grouping phase. Since logical grouping inherently involves identifying and organizing key features, an additional extraction step can introduce unnecessary overhead and distract the model from fully engaging its reasoning and planning abilities. Removing this step allows the model to implicitly perform feature identification while fully focusing on logical grouping to improve overall performance. 

In summary, our findings suggest that logical grouping is the most critical stage. Simplifying the CoT prompt to focus on this important stage, which aligns closely with the LLMs' strengths in reasoning and organization, can further enhance claim generation performance.

\begin{table*}[!t]
\centering
\footnotesize
\caption{Standard and CoT prompts (with variants) for patent claim generation.  }
\label{tab:prompt}
\begin{tabular}{|p{0.95\linewidth}|}
\toprule
\textbf{Normal Prompt:} \\
You are a patent expert. Given the patent description, your task is to write high-quality patent claims. \\
Output the claims ONLY.  \\
\midrule
\textbf{CoT Prompt: } \\
You are a patent expert. Given the patent description, your task is to write high-quality patent claims. \\
You should think step-by-step: \\
1. Identify the key technical features that define the invention and their importance ranking.  \\
2. Organize these features logically based on their functional relationships and dependencies.  \\
3. Write the claims using formal patent language to ensure clarity, consistency, and technical accuracy. \\
Output the claims ONLY. \\
\midrule
\textbf{Explicit CoT Prompt:} \\
You are a patent expert. Given the patent description, your task is to write high-quality patent claims. \\
You should think step-by-step: \\
1. Identify the key technical features that define the invention and their importance ranking.  \\
2. Organize these features logically based on their functional relationships and dependencies.  \\
3. Write the claims using formal patent language to ensure clarity, consistency, and technical accuracy. \\      
Example: \\
Input: \\
Patent description: \texttt{<<Description>>}  \\
Output: \\
\texttt{<|beginning of thinking|>} Ouput thinking \texttt{<|end of thinking|>} \\
\texttt{<|beginning of claims|>} Output claims \texttt{<|end of claims|>} \\
\midrule

\textbf{CoT Prompt without Extraction: } \\
You are a patent expert. Given the patent description, your task is to write high-quality patent claims. \\
You should think step-by-step: \\
1. Organize key technical features logically based on their functional relationships and dependencies.  \\
2. Write the claims using formal patent language to ensure clarity, consistency, and technical accuracy. \\
Output the claims ONLY. \\
\midrule

\textbf{CoT Prompt without Logical Grouping: } \\
You are a patent expert. Given the patent description, your task is to write high-quality patent claims. \\
You should think step-by-step: \\
1. Identify the key technical features that define the invention and their importance ranking. \\
2. Write the claims using formal patent language to ensure clarity, consistency, and technical accuracy. \\
Output the claims ONLY. \\
\midrule

\textbf{CoT Prompt without Formalization: } \\
You are a patent expert. Given the patent description, your task is to write high-quality patent claims. \\
You should think step-by-step: \\
1. Identify the key technical features that define the invention and their importance ranking.  \\
2. Organize key technical features logically based on their functional relationships and dependencies.\\
Output the claims ONLY. \\

\bottomrule
\end{tabular}

\end{table*}

\begin{table*}[ht]
\centering
\footnotesize
\caption{Example model outputs of patent EP3060168B1.}
\label{table:example_ouput}
\begin{tabular}{p{15.5cm}}
\toprule
\textbf{Gold claims} 1. A combination (10) comprising: a) an air tight container (12) having a water reservoir (14); and b) an injector (60) being retained in the container (12), wherein the injector is adapted to fold and inject an intraocular lens (16) into an eye of a patient; and c) the intraocular lens (16) being foldable and hydrophilic and disposed unfolded within the injector (60) so as not to be in direct contact with the water reservoir (14), characterized in that the container (12) comprises a tray (18) having an elongate compartment (20) wherein the tray (18) comprises a bottom wall (22) and side walls (26) wherein the water reservoir (14) is provided by blind holes (54) in interior surfaces of the bottom wall (22) and side walls (26) of the tray (18), wherein the water reservoir (14) retains between 0.5 and 3 ml of water, wherein the blind holes span a majority of the length and width of the tray (18). 
2. The combination (10) of claim 1 wherein the water reservoir (14) is provided by non-smooth textures (56) on the interior surfaces of the bottom wall (22) and side walls (26) of the tray (18) and wherein the water reservoir (14) comprises water retained within the surfaces with non- smooth textures (56). 
\\
\midrule

\textbf{GPT-4o (standard prompt)} 1. A packaging system for a foldable intraocular lens (IOL), comprising:    a. a substantially air-tight container having an interior volume;    b. a water reservoir configured to maintain a humidity level of at least about 90\% within the container;    c. an intraocular lens disposed within the container, wherein the lens is not immersed in liquid but is exposed to the humidity maintained by the water reservoir;    d. a cover sealing the container to minimize water loss through diffusion;    e. wherein the container is configured to allow condensation on the lens surfaces to maintain hydration adequate for foldability and delivery through an injection tube.  
2. The packaging system of claim 1, wherein the water reservoir comprises:    a. a water permeable pouch disposed within the container, the pouch containing water that is released over time to maintain the desired humidity level.  
3. The packaging system of claim 1, wherein the water reservoir comprises:    a. a water absorbing polymer capable of releasing water to the atmosphere within the container.  
4. The packaging system of claim 1, wherein the container includes:    a. void volumes separate from the location of the intraocular lens, configured to retain free water and minimize humidity gradients.  
5. The packaging system of claim 1, further comprising:    a. distribution enhancers within the container to facilitate even distribution of water vapor, selected from the group consisting of towels with wicking properties, pieces of plastic with blind holes, and textured surfaces on the interior of the container walls.  
6. The packaging system of claim 1, wherein the container is autoclavable and capable of withstanding temperatures of at least 121 degrees Celsius for sterilization purposes.  
7. The packaging system of claim 1, further comprising:    a. an injector disposed within the container, adapted to fold and inject the intraocular lens through a cylindrical injection tube having an inside diameter of 2 millimeters or less.  
8. The packaging system of claim 1, further comprising:    a. an autoclavable foil pouch enclosing the container, providing a second sterile barrier and minimizing water loss through diffusion. \\
\midrule

\textbf{GPT-4o (Implicit CoT)} 1. A packaging system for a foldable intraocular lens (IOL), comprising:    a. a substantially air-tight container having an interior volume, said container comprising a tray with a bottom wall, side walls, and an elongate top opening;    b. a cover sealing the elongate top opening of the container, said cover comprising a multilayered foil lid to minimize water loss through diffusion;    c. a water reservoir within the container, said reservoir providing an amount of water sufficient to moisture saturate the interior volume of the container at temperatures up to those required for steam sterilization, and to form small water droplets throughout the container;    d. wherein the water reservoir comprises a water permeable pouch disposed below the cover, said pouch designed to capture and retain water, releasing it as free water when atmospheric and other free water within the container is depleted over time by diffusion;    e. wherein the container is configured to maintain a humidity level of at least about 90\% within the interior volume to ensure adequate hydration of the IOL for foldability and insertion into an eye.  
2. The packaging system of claim 1, wherein the water reservoir further comprises distribution enhancers selected from the group consisting of:    a. a towel with wicking or sponge-like properties;    b. a piece of plastic with openings in the form of blind holes that span the surfaces of the tray;    c. a water absorbing polymer having a high water content that can be released to the atmosphere of the container, said polymer defining holes in which free water can reside.  
3. The packaging system of claim 1, wherein the tray is made from a thermoplastic material, such as polypropylene, and includes a circumferential rim with a raised outer lip surrounding a recessed inner band.  
4. The packaging system of claim 1, wherein the water reservoir is further provided by void volumes within the container, said void volumes being defined separate from the location of the intraocular lens within the container, and wherein the void volumes are maximized to minimize humidity gradients within the container.  
5. The packaging system of claim 1, wherein the water reservoir is further provided by blind holes in interior surfaces of the tray walls, said blind holes spanning a majority of the length and width of the tray, and/or by a non-smooth texture on the interior surfaces of the tray walls capable of retaining small amounts of free water.  
6. The packaging system of claim 1, wherein the combination of the container and water reservoir is autoclavable, capable of withstanding heating to 121 degrees Celsius or more for a period of a half hour or more.  
7. The packaging system of claim 1, further comprising an autoclavable foil pouch enclosing the container, said pouch minimizing water loss to diffusion, providing a second sterile barrier, and allowing the container to be introduced into a sterile field.  
8. The packaging system of claim 1, wherein the intraocular lens is disposed unfolded within an injector, said injector adapted to fold and inject the intraocular lens into the eye of a patient through a cylindrical injection tube having an inside diameter of 2 millimeters or less.
 \\
\midrule
\textit{Continue on next page} \\

\end{tabular}

\end{table*}

\begin{table*}[ht]
\centering
\footnotesize
\begin{tabular}{p{15.5cm}}

\textbf{GPT-4o (Explicit CoT)} \texttt{<|beginning of thinking|>} 
1. Key Technical Features and Importance Ranking:    - Packaging system for foldable intraocular lenses (IOLs) using water-saturated or nearly saturated air instead of immersion in liquid (High importance).    - Substantially air-tight container with minimal air volume to maintain hydration (High importance).    - Water reservoir system that maintains humidity and prevents water loss (Medium importance).    - Distribution enhancers for even water distribution within the container (Medium importance).    - Autoclavable packaging for sterilization (Medium importance).    - Use of a foil pouch for additional sealing and sterile barrier (Low importance).  
2. Logical Organization:    - Begin with the overall packaging system and its purpose.    - Describe the container's structure and its air-tight properties.    - Explain the water reservoir system and its function.    - Detail the distribution enhancers and their role in maintaining humidity.    - Include the autoclavable feature for sterilization.    - Conclude with the additional sealing provided by the foil pouch.  
3. Writing the Claims:    - Use formal patent language to clearly define each feature and its function.    - Ensure each claim is independent and builds logically on the previous one.    - Maintain clarity and technical accuracy throughout the claims. \texttt{<|end of thinking|>} 
\texttt{<|beginning of claims|>} 
1. A packaging system for foldable intraocular lenses, comprising:    a substantially air-tight container configured to house an intraocular lens, wherein the container maintains water-saturated or nearly saturated air to preserve the hydration of the lens without immersing it in liquid.  
2. The packaging system of claim 1, wherein the container comprises:    a tray with an elongate compartment bounded by a bottom wall, side walls, and an elongate top opening, the top opening being sealed by a multilayered foil lid to minimize water loss through diffusion.  
3. The packaging system of claim 2, further comprising:    a water reservoir within the container, the reservoir providing an amount of water sufficient to moisture saturate the interior volume of the container and form small water droplets throughout the container.  
4. The packaging system of claim 3, wherein the water reservoir includes:    a water permeable pouch disposed within the container, the pouch designed to capture and retain water, releasing it as needed to maintain humidity levels.  
5. The packaging system of claim 4, further comprising:    distribution enhancers within the container to facilitate even distribution of water, the enhancers including materials with wicking properties or structures with blind holes to retain free water.  
6. The packaging system of claim 1, wherein the container is autoclavable, capable of withstanding temperatures of at least 121 degrees Celsius for sterilization purposes.  
7. The packaging system of claim 1, further comprising:    an autoclavable foil pouch enclosing the container, providing an additional sterile barrier and minimizing water loss through diffusion.  8. The packaging system of claim 1, wherein the intraocular lens is disposed unfolded within an injector, the injector adapted to fold and inject the lens through a cylindrical injection tube with an inside diameter of 2 millimeters or less. 
\texttt{<|end of claims|>}
 \\

\bottomrule
\end{tabular}

\end{table*}

\end{document}